\documentclass[10pt,conference,final]{IEEEtran}
\IEEEoverridecommandlockouts
\usepackage{cite}
\usepackage{amsmath,amssymb,amsfonts}
\usepackage{algorithmic}
\usepackage{graphicx}
\usepackage{textcomp}
\usepackage{xcolor}
\def\BibTeX{{\rm B\kern-.05em{\sc i\kern-.025em b}\kern-.08em
    T\kern-.1667em\lower.7ex\hbox{E}\kern-.125emX}}

\usepackage{bm}         % better bold math
\usepackage{hyperref}   % for urls
\usepackage{tikz}

\newtheorem{assumption}{Assumption}

\usetikzlibrary{arrows}
\tikzstyle{var}=[circle,draw=black,fill=white,thin,minimum size=18pt,inner sep=0pt]
\tikzstyle{varh}=[circle,draw=gray,fill=white,thin,minimum size=18pt,inner sep=0pt,dashed]
\tikzstyle{vartarget}=[circle,draw=black,fill=lightgray,thin,minimum size=18pt,inner sep=0pt]
\tikzstyle{varintervention}=[rectangle,draw=black,fill=white,thin,minimum size=18pt,inner sep=0pt]
\tikzstyle{arr}=[->,>=stealth',draw=black,thick]
\tikzstyle{arrh}=[->,>=stealth',draw=gray,thick,dashed]
\tikzstyle{biarr}=[<->,>=stealth',draw=black,fill=black,thick]
\tikzstyle{biarrh}=[<->,>=stealth',draw=gray,fill=gray,thick]
\tikzstyle{arr}=[->,>=stealth',draw=black,thick]
\tikzstyle{arr-ch}=[o->,>=stealth',draw=black,thick]
\tikzstyle{arr-cc}=[o-o,>=stealth',draw=black,thick]
\tikzstyle{arr-hc}=[<-o,>=stealth',draw=black,thick]
\tikzstyle{arr-tc}=[-o,>=stealth',draw=black,thick]
\tikzstyle{arr-ct}=[o-,>=stealth',draw=black,thick]
\tikzstyle{arr-tt}=[-,>=stealth',draw=black,thick]

\renewcommand{\Pr}{\mathbb{P}}
\DeclareMathOperator*{\CI}{{\,\perp\mkern-12mu\perp\,}}
\DeclareMathOperator*{\nCI}{{\,\not\mkern-1mu\perp\mkern-12mu\perp\,}}

\newcommand{\an}[1]{\mathrm{an}\left(#1\right)}
\newcommand{\pa}[1]{\mathrm{pa}\left(#1\right)}

\renewcommand{\do}[1]{\rm{do}\left(#1\right)}

\begin{document}

\title{Boosting Local Causal Discovery in High-Dimensional Expression Data
\thanks{PV and JMM are supported by NWO VIDI grant 639.072.410.}}

\author{\IEEEauthorblockN{Philip Versteeg}
\IEEEauthorblockA{\textit{Informatics Institute} \\
\textit{University of Amsterdam}\\
Amsterdam, the Netherlands \\
p.j.j.p.versteeg@uva.nl}
\and
\IEEEauthorblockN{Joris M. Mooij}
\IEEEauthorblockA{\textit{Informatics Institute} \\
\textit{University of Amsterdam}\\
Amsterdam, the Netherlands \\
j.m.mooij@uva.nl}
}

\maketitle

\begin{abstract}
We study the performance of Local Causal Discovery (LCD)~\cite{Cooper1997}, a simple and efficient constraint-based method for causal discovery, in predicting causal effects in large-scale gene expression data. We construct practical estimators specific to the high-dimensional regime. Inspired by the ICP algorithm~\cite{Peters2016ICP}, we use an optional preselection method and two different statistical tests. Empirically, the resulting LCD estimator is seen to closely approach the accuracy of ICP, the state-of-the-art method, while it is algorithmically simpler and computationally more efficient. 
\end{abstract}


\section{Introduction}

% \IEEEpubid{978--1--7281--1867--3/19/\$31.00~\copyright~2019 IEEE }
\IEEEpubid{\begin{minipage}{\textwidth}\ \\ \\ \\ \\ \\ \\
\begin{center}978--1--7281--1867--3/19/\$31.00~\copyright~2019 IEEE.\end{center} 
 Personal use of this material is permitted.  Permission from IEEE must be obtained for all other uses, in any current or future media, including reprinting/republishing this material for advertising or promotional purposes, creating new collective works, for resale or redistribution to servers or lists, or reuse of any copyrighted component of this work in other works.
 \end{minipage}}
\IEEEpubidadjcol

% Intro
One of the main goals of empiricism is uncovering relationships that underlie the data at hand. Scientific questions often aim for the estimation of causal quantities instead of statistical associations. Causal models can predict the effects of perturbations (interventions) to the system, allowing one to reason about previously  unseen experiments that are too costly, difficult or unethical to perform. 

% Causal inference
One approach to causal inference is constraint-based, where statistical evidence in the data on multiple variables at a time is combined to limit the search space of possible causal effects. The PC and FCI algorithms~\cite{sgs00} ingeniously combine independence testing under certain assumptions to infer a set of possible causal graphs. These methods are sound and complete under their respective assumptions, but require a (possibly very slow) global search when the variable set is large and relations are non-sparse. Furthermore, they are traditionally only applied to observational data, not making use of any perturbation data that may be available. 

% Main point!
The LCD algorithm~\cite{Cooper1997}, one of the simplest constraint-based causal discovery methods, uses background information, that can be derived from experiments, in addition to (in)dependencies to predict that one variable causes another for subsets of three variables locally. The method is algorithmically simple and generic such that many practical variants are possible. Specifically, a conservative version of LCD has been used to infer protein signaling networks from mass cytometry data~\cite{triantafillou2017predicting}. Another estimator related to LCD is Trigger~\cite{chen2007harnessing}. 

The more recently introduced ICP algorithm~\cite{Peters2016ICP} requires data from multiple contexts. Under certain assumptions, it predicts causal effects by exploiting that certain conditional distributions remain invariant under changes of the context. In~\cite{Meinshausen2016Pnas} ICP predictions have been assessed with real-world gene expression data~\cite{Kemmeren2014}, and results were shown to be significant when compared to an internal data-driven ground-truth and several externally sourced gold-standards. 

% Main contribution
In this work, we focus on the statistical aspects of LCD and apply it to predict knock-out effects in large-scale gene expression data~\cite{Kemmeren2014}. Inspired by the implementation of the ICP estimator, we construct several practical versions of LCD that are well-suited to this high-dimensional task. In particular, we apply $L_2$-boosting, a fast gradient boosting regression method for variable selection suitable to the high-dimensional setting~\cite{schapire1998boosting},~\cite{buhlmann2003boosting}, to limit the search space of LCD\@. We also examine the effect of using different statistical tests in the LCD estimator. 

\IEEEpubidadjcol

% Problem setup
The experiments are performed on train-test splits of gene expression data~\cite{Kemmeren2014} containing ``pure observations'' and knock-outs. Here one part of the knock-out data serves as the background information that can be used in prediction methods, while the remaining data are used to construct a ground-truth set. We compare the performance of the LCD estimators to ICP, the state-of-the-art method for this dataset~\cite{Meinshausen2016Pnas} that exploits the perturbed data in a more sophisticated approach. We find that implementing the preselection procedure with LCD results in cause-effect predictions that evaluate comparably to ICP, while LCD is computationally more efficient. The choice of the statistical test was not seen to affect results. 

% outline
We start by concisely introducing causal models and causal inference from both observational and interventional data. Then we discuss the main methods in this work, LCD and ICP, and their estimators. We outline the experimental approach and provide results for experiments on gene expression data, before we close with conclusions.

\section{Preliminaries}\label{sec:preliminaries}

Here we shortly introduce causal graphical modeling and methods for causal inference from observational and interventional data.

\subsection{Graphical Causal Models}

% causal discovery
We consider a set of $p$ random variables $\bm{X} = \{X_1, X_2, \dots, X_p\}$, for which the observed causal relationships are modeled as edges between nodes (variables) in a directed mixed graph (DMG).
% and a given Structural Causal Model $\mathcal{M}$.  
Here, a directed edge $X \rightarrow Y$ between $X,Y \in \bm{X}$ corresponds to a \emph{direct causal effect} relative to $\bm{X}$. An \emph{ancestral causal effect}, typically plainly referred to as a \emph{causal effect}, then corresponds to the existence of a sequence of direct causal effects $X \rightarrow W \rightarrow V \rightarrow \dots \rightarrow Y$. The set of direct causes of a target variable $Y$ is denoted as its \emph{parents} (or $\pa{Y}$), while the set of its ancestral causes is referred to as its \emph{ancestors} (or $\mathrm{anc}(Y)$). 
% confounders
When unmeasured confounding is present (in the case without \emph{causal sufficiency}), the causal graph is a directed mixed graph (DMG) that contains bidirected edges between variables, indicating latent unmeasured variables that causally affect pairs of measured variables.

% SCM
A common way of modeling observational and interventional data related to a causal graph is with a Structural Causal Model (SCM)~\cite{pearl2009causality}.\footnote{Technically we consider the class of Simple SCMs here (see \cite{mmc2018jci}), for which, relative to a given SCM, the formal definitions of (direct) causal effects coincide with the more informal notations here.} 
In essence, an SCM specifies a single equation for each \emph{endogenous} variable $X_i$,
\begin{equation}\label{eqn:scm}
    X_i = f_i \left(\mathrm{pa}(X_i), N_i \right)\mathrm{,}
\end{equation}
where $f_i$ is a function of the parent set $\pa{Y}$ and an \emph{exogenous} noise variable $N_i$. The solutions of these structural equations, together with the distributions of the exogenous noise terms, specify the joint observational distribution $\Pr\left(\bm{X}\right)$. Perfect interventions are modeled by replacing the right-hand side of~\eqref{eqn:scm} for all targeted variables $i \in \bm{I} \subseteq \bm{X}$ by the corresponding fixed value $\xi_i$, inducing the corresponding \emph{interventional distribution} $\Pr\left(\bm{X} \,|\, \mathrm{do}\left(\bm{X}_{\bm{I}} = \bm{\xi}_{\bm{I}}\right)\right)$. We refer the reader to~\cite{pearl2009causality} and~\cite{bongers2016theoretical} for detailed treatments on SCMs.

\subsection{Constraint-based Causal Discovery}
The \emph{Markov assumption} (see e.g.~\cite{pearl2009causality},~\cite{forre2017markov} for an overview) relates statistical properties of a causal graph %induced by a SCM 
to independence constraints in observed data. Missing edges in the causal graph imply (conditional) independencies of the form $X \CI  Y | \bm{Z}$, for $\bm{Z} \subset \bm{X}$, through the $d$-separation criterion (see e.g.~\cite{sgs00,pearl2009causality}) in the acyclic case.\footnote{Recent advances have non-trivially extended $d$-separation to cyclic models by introducing $\sigma$-separation~\cite{forre2017markov}. It can be shown that the LCD and ICP (reformulated in JCI-1 terms) methods are valid  when cycles are present for Simple SCMs by using this separation criterion under certain modeling assumptions~\cite{mmc2018jci}. As it is not the main focus of this work, we will treat methods mainly in terms of acyclic models and $d$-separation.} 
Under the \emph{faithfulness assumption} the reverse implication holds, such that \emph{all} independencies present in the data correspond to pairs of separated variables. 

% algos
Algorithms in a class of \emph{constraint-based} causal discovery methods build on this principle, learning (direct) causal effects from independence statements that are derived from observational data. The PC algorithm combines independencies between all pairs of variables in a clever search to recover (as much of) an acyclic causal graph under causal sufficiency, whereas the FCI algorithm extends this strategy to cases where confounding and selection bias are present~\cite{sgs00}. These methods can be computationally prohibitive for systems with a large number of variables, as the search can require a large number of pairwise tests. Local strategies such as Y-Structures~\cite{mooij2015Ystructures} and LCD~\cite{Cooper1997} (see Sec.~\ref{sec:methods_lcd}) trade off completeness for the ability to scale to a large number of variables.

\subsection{Interventional Data and Joint Causal Inference}

% interventions
In almost all practical situations, only a subset of the causal effects is recoverable from independence statements derived from purely observational data. Even in the infinite sample limit, observational constraint-based causal discovery is in general feasible up to an equivalence class that represents multiple indistinguishable graphs. The identifiability of causal effects can be increased by the inclusion of interventional samples.

% jci
The recently introduced Joint Causal Inference (JCI)~\cite{mmc2018jci} is a framework for incorporating various types of experimental data jointly in causal learning problems. 
Perfect interventions (and other manipulations) are jointly modeled through multiple \emph{context variables}, in addition to the regular endogenous \emph{system variables}, and a set of underlying assumptions.
For example, samples from multiple datasets can be combined by including a single context variable to the dataset, resulting in an extra integer column in the dataset that encodes the originating dataset of each sample.
The main idea of JCI is then to jointly learn over the meta-system of combined system and context variables, while encoding a set of required assumptions on context variables as background knowledge. 

% assumptions
The JCI assumptions that are relevant for methods in this work are given as follows, where $\bm{C}$ and 
 $\bm{X}$ now denote sets of context and system variables respectively.
\setcounter{assumption}{-1} 
\begin{assumption}(\emph{Joint SCM}) 
    The data-generating process underlying $\bm{C}$ and $\bm{X}$ is modeled by a single (simple) SCM\@.
\end{assumption}
\begin{assumption}
    (\emph{Exogenity}) No system variable $X \in \bm{X}$ is an ancestor of any context variable $C \in \bm{C}$.
\end{assumption}
This combined modeling assumption is labeled as the JCI-1. We refer the reader to~\cite{mmc2018jci} for additional JCI assumptions related to context models that are outside the scope of this work. 

% Example
As a simple example of a JCI-1 model, consider the Randomized Controlled Trial (RCT), where the causal effect of a randomized treatment on a single outcome variable $Y$ is studied. Here the context $C$ indicates the treated population ($C=1$) and the control group ($C=0$), expressing the belief that outcome of the treatment does not effect the treatment itself. Under these conditions, i.e.~assuming JCI-1, and when there is no confounding and selection bias, a statistical dependence between $C$ and $Y$ implies a causal effect of $C$ on $Y$.

\section{Methods}\label{sec:methods}

We first describe the two causal prediction methods that are central to this work in the terminology of JCI\@. We discuss practical estimators for each algorithm specifically.

\subsection{Invariant Causal Prediction}\label{sec:methods_icp}

% intro
The Invariant Causal Prediction (ICP)~\cite{Peters2016ICP} algorithm predicts a conservative subset of the true parent set for a given target variable $Y$ by identifying conditional distributions that remain invariant under changes of a single context variable $C$, also referred to as the \emph{environment}. In the original formulation, ICP assumes that there is no causal effect from $C$ to $Y$ in addition to causal sufficiency and  acyclicity, but it remarkably does not require the faithfulness assumption. One of the main statistical aspects is its ability to set a confidence level for which subsets are accepted.

% Jci-ICP
In~\cite{mmc2018jci} it is shown that ICP can be reformulated in terms of JCI-1 when faithfulness is additionally assumed, and the interpretation proposed there also allows for cycles to be present. ICP then outputs a subset of the true ancestors of $Y$ instead of a subset of the parent set. 

We now describe several statistical aspects of a practical ICP estimator, which is one of possible choices given the generic ICP algorithm.

\paragraph*{Statistical Test}\label{sec:icp_stat_test}
The original ICP variant~\cite{Peters2016ICP} tests at confidence level $\alpha$ if a linear SCM exhibits invariant model parameters in realizations $c$ of $C$ for a given target effect $Y$. 
The authors of~\cite{Peters2016ICP} propose two practical estimators, a regression based test and an approximate test. For the first we refer to Sec.~3.1.1 in~\cite{Peters2016ICP}, while the second test, which we label a \emph{mean-variance} test, is constructed as follows.

% mv test 
For each potential parent set, i.e.~each subset of $\bm{X} \setminus \{Y\}$, the mean of the residuals of a linear regression in each context $c \in C$ is compared to the mean of the linear regression residuals in all others contexts $C \setminus \{c\}$ with a $t$-test.
The resulting $p$-values are combined over all contexts with a Bonferroni correction. Analogously, an $F$-test is used to test differences in variances of the residuals across contexts and combined in the same way. The final $p$-value of the potential parent set is the minimal of the two values, and the returned parent set is chosen as the intersection of all potential parents that are not rejected at level $\alpha$. For example, the test returns both $X_1$ and $X_2$ as parents of $Y$ in the linear model in Fig.~\ref{fig:icp}. 

\paragraph*{Preselection}
In the worst case, ICP scales exponentially in the number of system variables $p$ by examining all subsets of variables as potential causes. To circumvent this problem,~\cite{Peters2016ICP} suggests limiting this set of potential parents for each target $Y$ through a preselection method. Their proposal is to take a number of top ranking non-zero coefficients from a Lasso regression~\cite{tibs96} or $L_2$-boosting~\cite{schapire1998boosting} regression, where the latter  essentially is the repeated application of least square regression fitting of residuals.

\paragraph*{ICP Estimator}
For each of the above options, we pick the default setting used in the \texttt{R} package \texttt{InvariantCausalPrediction} for the high-dimensional setting encountered here. Thus, the practical \textsc{icp} estimator uses the mean-variance test and the $L_2$-boosting preselection.\footnote{We use small capitalization to label estimators.} 

\subsection{Local Causal Discovery}\label{sec:methods_lcd}

% Intro
Local Causal Discovery (LCD)~\cite{Cooper1997} is a constraint-based causal discovery method, combining three (in)dependence tests with causal background knowledge. It requires that the data includes (at least) one context variable $C$ subject to JCI-1 assumptions and that selection bias is not present in the samples. LCD does not assume causal sufficiency and cycles may be present in the modern formulation~\cite{mmc2018jci}.

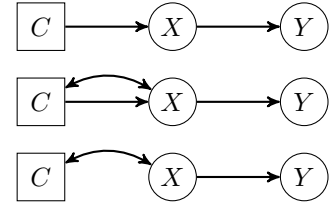
\begin{figure}[htbp]
\centering
\centerline{
    \begin{tikzpicture}
        % graph 1
        \node[var, rectangle] (C1) at (-1.75,1) {$C$};
        \node[var] (X1) at (0,1) {$X$};
        \node[var] (Y1) at (1.75,1) {$Y$};
        \draw[arr] (C1) edge (X1);
        \draw[arr] (X1) edge (Y1);
        % graph 2
        \node[var, rectangle] (C) at (-1.75,0) {$C$};
        \node[var] (X) at (0,0) {$X$};
        \node[var] (Y) at (1.75,0) {$Y$};
        \draw[arr] (C) edge (X);
        \draw[arr] (X) edge (Y);
        \draw[biarr, bend left] (C) edge (X);
        % graph 3
        \node[var, rectangle] (C2) at (-1.75,-1) {$C$};
        \node[var] (X2) at (0,-1) {$X$};
        \node[var] (Y2) at (1.75,-1) {$Y$};
        \draw[arr] (X2) edge (Y2);
        \draw[biarr, bend left] (C2) edge (X2);
    \end{tikzpicture}}
\caption{All three-variable DMGs with context $C$ where LCD predicts that $X \in \pa{Y}$ relative to $\{C,X,Y\}$.}\label{fig:lcd}
\end{figure}

% Workings
LCD tests if the following constraints hold for a given triplet of variables $\{C,X,Y\}$:
\begin{align}\begin{split}\label{eqn:lcd}
    C \CI & Y | X \\
    C \nCI & X \\ 
    X \nCI & Y
\end{split}\end{align}
where $X,Y \in \bm{X}$ and $\nCI$ denotes statistical dependence. It then follows that $X$ is a direct cause of $Y$ relative to $\{C,X,Y\}$, $Y$ is not a direct cause of $X$ and that no confounding is present between $X$ and $Y$. One straightforward proof is to enumerate all possible DMGs for three variables where these (in)dependencies and the JCI-1 assumption holds, resulting in the three DMGs in Fig.~\ref{fig:lcd}. 

% Local
As LCD holds under confounding, it can directly be applied to systems with more than three variables by iterating over (all) subsets of triples, each time marginalizing all other variables and testing for the LCD constraints in~\eqref{eqn:scm}. In that case, the LCD prediction is an \emph{ancestral} causal effect, rather than a \emph{direct} causal effect, relative to all system and context variables. 

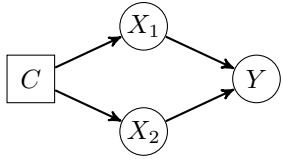
\begin{figure}[htbp]
\centering
\centerline{
    \begin{tikzpicture}
        \node[var, rectangle] (C) at (-1.5,0) {$C$};
        \node[var] (X1) at (0,0.7) {$X_1$};
        \node[var] (X2) at (0,-0.7) {$X_2$};
        \node[var] (Y) at (1.5,0) {$Y$};
        \draw[arr] (C) edge (X1);
        \draw[arr] (C) edge (X2);
        \draw[arr] (X1) edge (Y);
        \draw[arr] (X2) edge (Y);
    \end{tikzpicture}}
\caption{Example of a DMG with context $C$ where LCD does not make a prediction, but where ICP returns $\{X_1, X_2\}$ as parent set of $Y$.}\label{fig:icp}
\end{figure}

% Difference LCD to ICP
Compared to ICP, the LCD algorithm exhibits less power in predicting causal effects. ICP predicts the causal effect $X \in \pa{Y}$ in each of the LCD patterns in Fig.~\ref{fig:lcd}, but it can additionally infer multiple parents for a single effect when configured as in Fig.~\ref{fig:icp}. The latter pattern is omitted by LCD, as marginalizing over $X_1$ (or $X_2$) leads to a dependency between $C$ and $Y$ even when conditioning on $X_2$ (or $X_1$). Note however that ICP is less efficient, as it potentially searches over all possible subsets of $\pa{Y}$, whereas LCD has a computational complexity of $\mathcal{O}\left(p^3\right)$.

% \subsubsection*{LCD Estimators}
We introduce four practical LCD estimators here that each implement the generic LCD algorithm described above. Each one of these estimators is a combination of the following two options.

\paragraph*{Statistical Test}
We consider two options for testing the (in)dependencies in~\eqref{eqn:lcd} at confidence level $\alpha$. Following common practice in constraint-based causal discovery literature, we accept independence whenever $p \geq \alpha$. 

We can use the same combined mean-variance test as used by the \textsc{icp} estimator in Sec.~\ref{sec:methods_icp} to find potential parent sets. 

Alternatively, we use a (partial) correlation test, one of the simplest (conditional) independence test that assumes a Gaussian distribution and which is used extensively in constraint-based causal discovery. The $p$-value for the null hypothesis of independence is computed using the Student's $t$ transform of the (partial) correlation coefficient. 

\paragraph*{Preselection}

As LCD scales reasonably well with the number of variables $p$, it is generally possible to compute all possible LCD triples for large $p$ with a fast test.

Alternatively, we use $L_2$-boosting as a preselection method inspired by the ICP estimator. Then we only test for LCD triples $\{C,X,Y\}$ if $X$ is found is in the variable selection for $Y$ for suitable tuning of the boosting method.

\paragraph*{LCD Estimators}
This results in the following practical LCD estimators.
\begin{itemize}
    \item \textsc{lcd}: no preselection and the partial correlation test.
    \item \textsc{lcd-mv}: no preselection and the mean-variance test.
    \item \textsc{lcd-bst}: $L_2$ boosting as preselection and the partial correlation test.
    \item \textsc{lcd-bst-mv}: $L_2$ boosting as preselection and the mean-variance test.
\end{itemize}

\subsection{Boosting Baseline}\label{sec:methods_bst}

An interesting non-causal baseline is obtained by simply computing the $L_2$-boosting preselection method for each target variable $Y$ and naively labeling the set of selected variables (that are predictive for $Y$) as the causes of $Y$.

\section{Experiments}

We apply the LCD and ICP estimators on train-test splits of gene expression data, that consists of both observational and interventional data labeled by a JCI context variable. We define a target ground-truth score, derived from part of the ``true'' interventional measurement and a few observational samples, and assess the performance of each predictor against this score.

\subsection{Experimental Setup}

\begin{figure}[htbp]
\centering
\centerline{
\includegraphics[width=0.24\textwidth]{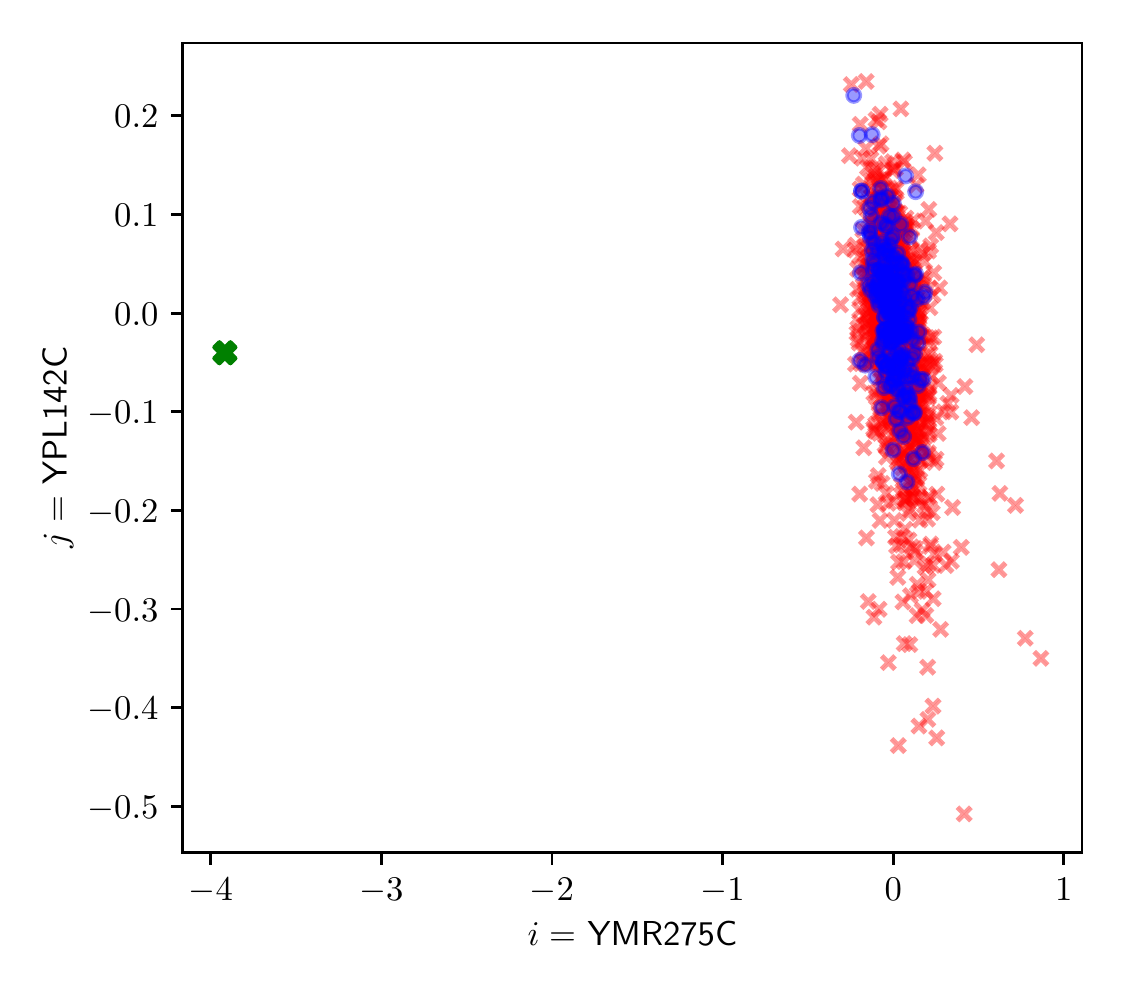}
\includegraphics[width=0.24\textwidth]{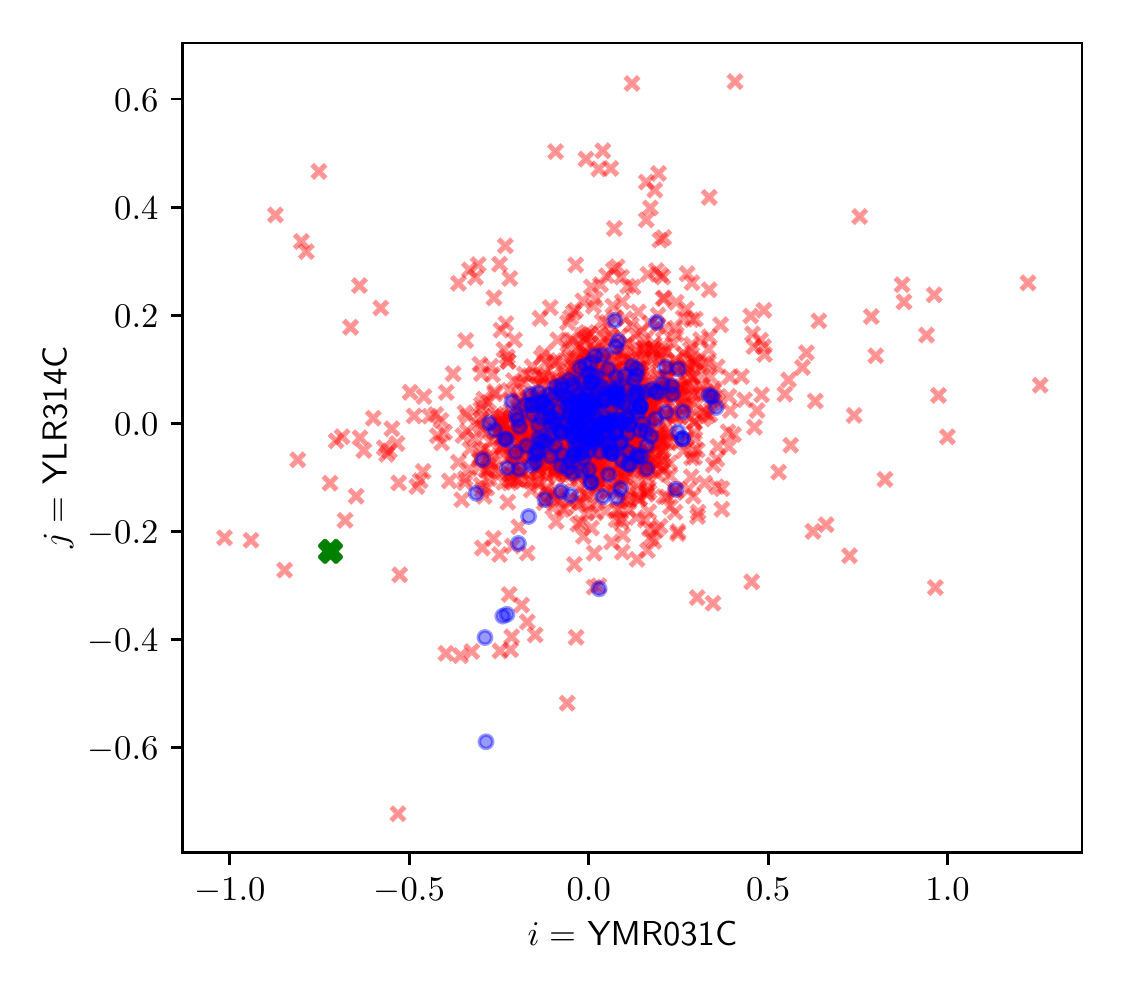}
}
\caption{Examples of pairwise gene expressions $(i,j)$ from the~\cite{Kemmeren2014} dataset. Observational and interventional samples are shown in blue and red, while the knock-out of the gene $i$, which expression is displayed on the horizontal axes (e.g.~\textit{YMR275C} and \textit{YMR031C}), is indicated by the green cross. 
}\label{fig:genedata}
\end{figure}

\subsubsection*{Data}

We use data from microarray experiments from~\cite{Kemmeren2014}, where, after preprocessing, $262$ \emph{observational} samples are available for each of $6170$ genes.\footnote{Original and preprocessed data files are available at \url{http://deleteome.holstegelab.nl/downloads_causal_inference.php}.} Additionally, in $1479$ knock-out experiments a unique single gene has been externally disabled (intervened) and the expression level is measured once for all $6170$ genes. In Fig.~\ref{fig:genedata} we show how these \emph{interventional} samples differ from observational data for a few example gene pairs.

\subsubsection*{Train-Test Split}
In a train-test setup, we split the dataset in five equally sized and disjoint partitions, such that each one consist of $1/5$th of the observational samples and $1/5$th of the interventional experiments. Each of these parts is considered once as a test set, in which the knock-out data are used to construct the ``true'' causal effect of interventions disjoint from the training set. The remaining four parts are then merged in a training set, that therefore contains all interventional samples that are \emph{not} the target ``true'' causal effect. 
% JCI
The training data are prepared as a joint JCI dataset with a single context variable and $6170$ system variables, where all observations are gathered in one context labeled $C=0$ and all interventions in another labeled $C=1$.

% test
\subsubsection*{Ground-Truth Definition}
The interventional samples, where in each one a single knock-out (i.e.~a perfect intervention) has been performed, can straightforwardly be used to test for the existence of (ancestral) causal effects for that knock-out. A significant difference in the interventional distribution $\Pr{\left(Y | \mathrm{do} \left(X\right)\right)}$ from the observational distribution $\Pr{\left(Y \right)}$ implies that $X \in \an{Y}$, where all other variables $\bm{X} \setminus \{X,Y\}$ are marginalized. This allows us to construct a ground-truth score from the interventional data that scores causal effect as follows.

Let the expression of gene $j$ measured under intervention $i$ be denoted as $X_{j;i}$. This single interventional sample represents the raw expression level in the units of measure of gene $j$, independent of its natural observational state. We standardize it with respect to the observational expression levels of this gene, resulting in the following score:
\begin{equation}\label{eqn:gt}
    S_{ij}^{\mathrm{std}} = \frac{| X_{j;i} - \mu_j |}{\sigma_j}\mathrm{,}
\end{equation}
where $\sigma_j$ and $\mu_j$ are the empirical mean and standard deviation of the expression level of gene $j$ in the observational data. 
We compute this score for all combinations of interventions and all genes that are available in each test set and merge the score afterwards over all pairs. The result is a total of $9119260$ scored pairs that are available in the ground-truth, for $1479$ unique interventions on $6170$ genes, where we have excluded self-effects.

\subsubsection*{Evaluation} 
For a given training set, predictions for each causal method are computed only for those intervention-gene pairs $(i,j)$ for which the intervention is available in the respective test set. All prediction scores are aggregated afterwards over all partitions. The continuous ground truth score $S_{ij}^{\mathrm{std}}$ is thresholded at a desired prevalence level to produce a ground-truth set of ``true'' pairs. The pairs that rank highest according to each of the prediction methods are then compared with this set in a ROC curve, where the recall, or true positive rate (TPR), is set out against the sensitivity, or false positive rate (FPR).

\begin{figure*}[htbp]
\centering
\centerline{
\includegraphics[clip, trim=0pc 0pc 2pc 2pc, width=0.33\textwidth]{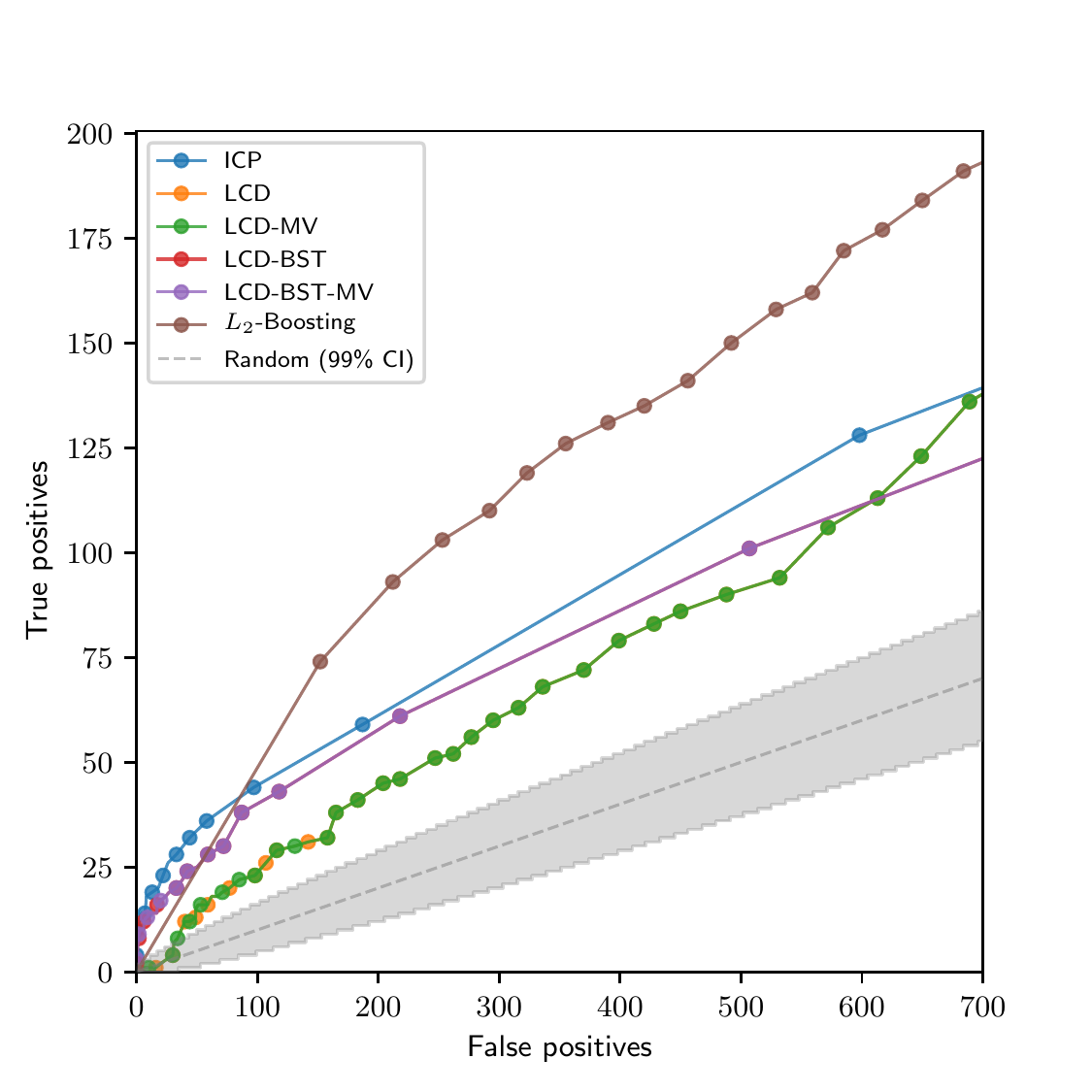}
\includegraphics[clip, trim=0pc 0pc 2pc 2pc, width=0.33\textwidth]{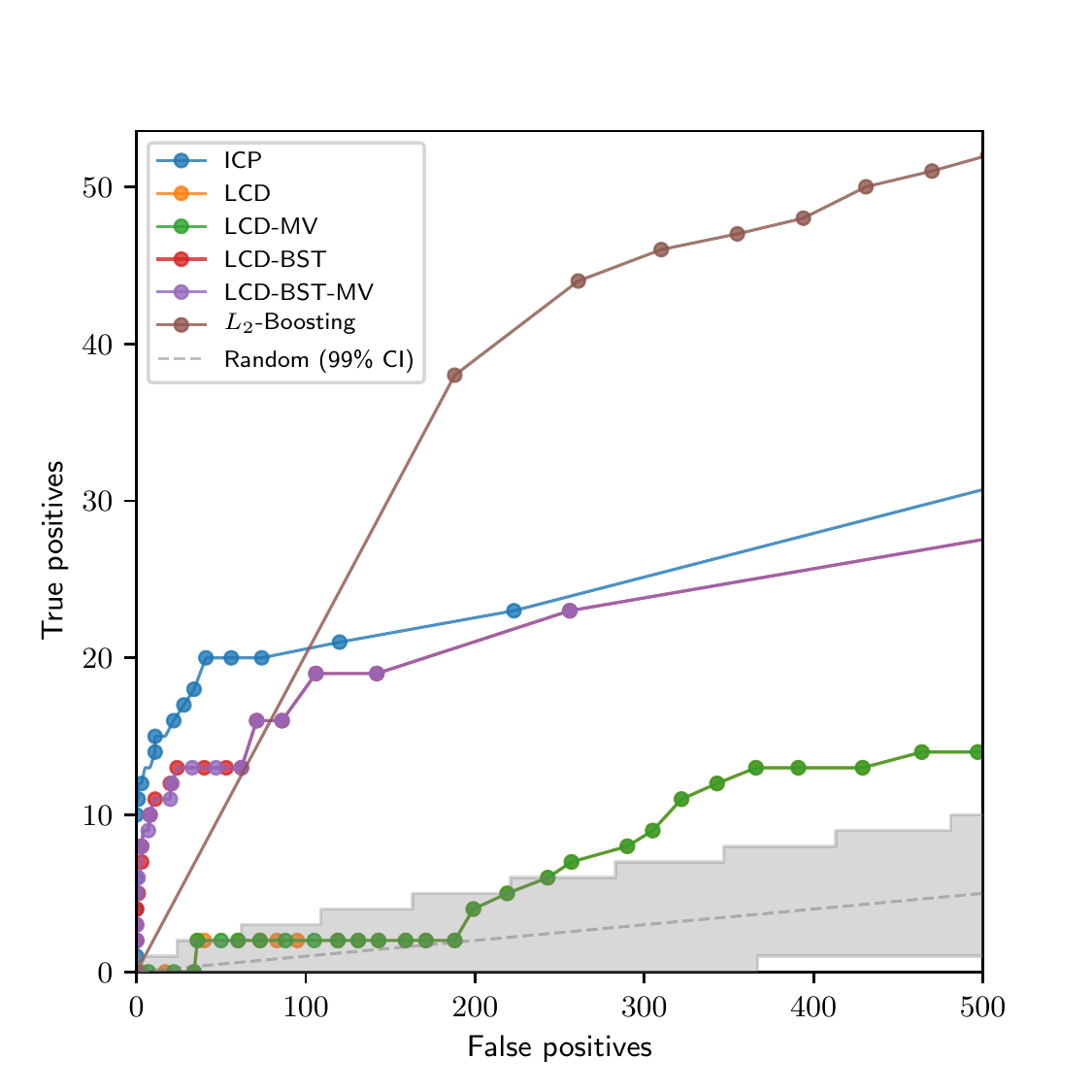}
\includegraphics[clip, trim=0pc 0pc 2pc 2pc, width=0.33\textwidth]{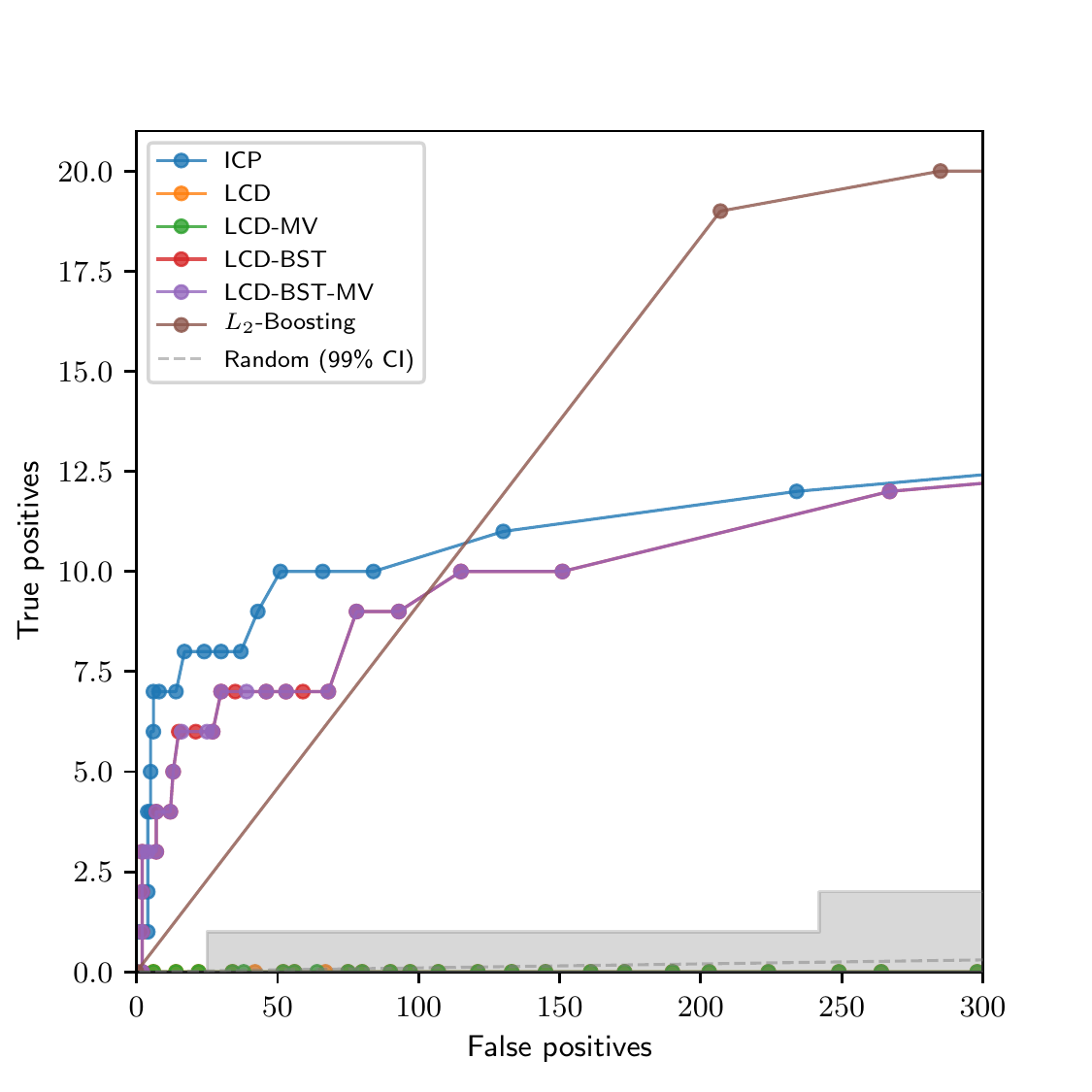}}
\caption{ROC curves for the top predictions of \textsc{icp}, four \textsc{lcd} estimators and the $L_2$-boosting baseline in a $5$-fold train-test split of the gene expression data. Curves for \textsc{lcd} almost completely match \textsc{lcd-mv}
, and similarly for \textsc{lcd-bst} and \textsc{lcd-bst-mv}. From left to right, the ground-truth is composed of the $10$\%, $1$\% and $0.1$\% of the top scoring pairs according to $S_{ij}^{\mathrm{std}}$ respectively. Random guessing with a $99
$\% confidence interval is shown in gray.}
\label{fig:kem_roc}
\end{figure*}

\subsection{Method Implementation}

% ICP
We used the \texttt{InvariantCausalPrediction}~\cite{Peters2016ICP} \texttt{R} package to compute ICP predictions (Sec.~\ref{sec:methods_icp}). We closely follow the approach of~\cite{Meinshausen2016Pnas}: the confidence level for the ICP estimator is set at $0.01$ and \texttt{stopIfEmpty} is set to \texttt{TRUE} for faster computation, while all other settings are left to the defaults. This implies that $L_2$-boosting is used as preselection method, which calls the \texttt{glmboost} routine from the \texttt{MBoost} \texttt{R} package. Here the tuning parameter is set at the ICP default, such that a maximum of $8$ variables are selected as potential parents for each target variable, while all other parameters are left at default options. Note that this variable selection is also computed over the joint data, which includes all observational and interventional samples.

% LCD
For the LCD estimators (Sec.~\ref{sec:methods_lcd}) we use the same settings as ICP whenever applicable: a confidence level of $0.01$ for each of the statistical tests is used, and the same construction of the $L_2$-boosting is applied in for preselection in \textsc{lcd-bst} and \textsc{lcd-bst-mv}.

% L_2 boosting
The non-causal $L_2$-boosting baseline (Sec.~\ref{sec:methods_bst}) uses the same preselection construction as described above for LCD and ICP\@.

% Bootstrap
All prediction methods are trained over $100$ random subsamples of the training data to increase the stability of predictions in the high-dimensional setup~\cite{pbvdg11}. In the subsampling procedure, the joint observational and interventional data are sampled uniformly at a fraction of $0.5$ without replacement. The resulting final stabilized~\cite{mebu10} estimator for each prediction method that is used is then given as the number of pairs predicted across all subsamples.

% Trigger
Lastly we attempted to compare with the Trigger estimator~\cite{chen2007harnessing}, which assumes that a natural (Mendelian) randomization of the genetic information underlies the data and then combines a sequence of likelihood ratio tests. We found however that the practical estimator in the \texttt{Trigger} \texttt{R} package could not be computed successfully on a pooling of the knockout expression data with context variables.

\subsection{Results}

% describe results
The results of the experiments are shown in Fig.~\ref{fig:kem_roc}, where the ICP, LCD and the baseline estimators are compared to the top scoring pairs according to the ground-truth score~\eqref{eqn:gt}. ROC curves are shown at three different levels of prevalence, where in each figure we vary the number of true effects that are in the ground-truth set.

% bst
Surprisingly, the $L_2$-boosting baseline shows a very effective selection of the first few hundred pairs in the ground-truth set. Nonetheless, it is still outperformed by ICP and LCD at low prevalence, as expected.

% icp
We find that \textsc{ICP} outperforms other methods for the top ranking predictions for this particular definition of the ground-truth, where a large portion of the highest scoring predictions (i.e.~the pairs that are predicted most frequently in all random subsamples) matches with the top $10$\%, $1$\% and $0.1$\% pairs in the ground-truth.
This reaffirms the findings of~\cite{Meinshausen2016Pnas}, where a different ground-truth definition is used. 

% LCD
The simplest LCD-type estimator, \textsc{lcd}, shows better-than-random results at the $10$\% prevalence, but this is reduced to random guessing levels for the harder prediction tasks at lower prevalences. The same applies to \textsc{lcd-mv}, which uses the mean-variance test. Applying a $L_2$-boosting preselection procedure has a large effect, as can be seen in the performance of \textsc{lcd-bst}. This method shows a larger recall at the same level of FPR for each of the settings, approaching the \textsc{icp} result in all cases. When we compare \textsc{lcd-bst-mv} to \textsc{lcd-bst}, it is seen that the choice of using either the partial correlation test or the more complex mean-variance test has almost no effect on the evaluation with respect to the top scoring pairs. 

\begin{table}[htbp]
\caption{Computation Times (CPU hours)}
\begin{center}
\begin{tabular}{|c|c|cccc|}
\hline
\textsc{$L_2$-boosting} & \textsc{icp} & \textsc{lcd} & \textsc{lcd-mv} & \textsc{lcd-bst} & \textsc{lcd-bst-mv} \\
\hline
 118 & 585 & 17 & 1936 & 133 & 172 \\ 
\hline
\end{tabular}
\label{tab:comp_times}
\end{center}
\end{table} 

% computation 
Computation times are shown in Tab.~\ref{tab:comp_times} for each method, where we report the total time required for computing all pairwise predictions over all $100$ subsamples.\footnote{The experiments were computed in an embarrassingly parallel setup with CPUs with comparable qualities, where each subsample of the data was randomly computed on either an Intel Xeon E5--2680, Intel Xeon E5--2680 or Intel Xeon Gold--5118 processor.} The mean-variance test is two orders of magnitude slower in \textsc{lcd-mv} than simply running \textsc{lcd} with partial correlations, while having little to no effect on outcomes (Fig.~\ref{fig:kem_roc}). Preselecting results with the $L_2$-boosting leads to a sparse enough set of possible parents such that a speedup is seen for the \textsc{lcd-bst-mv} estimator, in addition to increased precision. Applying the $L_2$-boosting preselection to the simplest LCD implementation, \textsc{lcd}, which uses the fast partial correlations test, results in the \textsc{lcd-bst} estimator, which is an order of magnitude slower but with better sensitivity.

% final
We also find that \textsc{icp} is several times slower than \textsc{lcd-bst-mv}, the estimator that uses the same preselection and independence test as \textsc{icp}. Notably, \textsc{icp} is even less efficient when compared to \textsc{lcd-bst}, while both \textsc{lcd-bst} and \textsc{lcd-bst-mv} are very similar in performance to \textsc{icp}.

\section{Conclusion}

% summarized result 
Inspired by features of the ICP estimator, we implemented several different practical estimators of LCD, a simple constraint-based causal discovery method.
We have shown that LCD predictions, computed on top of an $L_2$-boosting procedure, can successfully predict the effects of unseen gene knock-outs in large-scale expression data. The estimator closely approximates the empirical performance of the state-of-the-art ICP, while it is algorithmically simpler and computationally more efficient. We also found that the predicted causal effects from LCD in this setting are robust under the choice of the independence test that is used. Surprisingly, a large part of the good performance is already explained by the $L_2$-boosting preprocessing that these algorithms apply before causal considerations come into play. 

% outlook
As we have seen, statistical aspects of causal discovery methods have a large impact on empirical performance. Nonetheless, they are largely underexplored for many practical algorithms. This leaves open the possibility for future work.

\section*{Acknowledgment}
The authors thank Age Smilde and an anonymous reviewer for valuable comments on the manuscript.

% \section*{References}
\bibliographystyle{IEEEtranS}
\bibliography{IEEEabrv,references.bib}

\end{document}